\documentclass[10pt,twocolumn,letterpaper]{article}

\usepackage{wacv}              
\makeatletter
\def\ps@plain{%
  \let\@oddhead\@empty
  \let\@evenhead\@empty
  \let\@oddfoot\@empty
  \let\@evenfoot\@empty
}
\makeatother

\usepackage{graphicx}
\usepackage{amsmath}
\usepackage{amssymb}
\usepackage{booktabs}
\usepackage{caption}
\usepackage{color,soul} 
\soulregister\ref{1}
\soulregister\eqref{1}
\soulregister\cite{1}

\usepackage{enumitem}

\usepackage[pagebackref,breaklinks,colorlinks]{hyperref}

\usepackage[capitalize]{cleveref}
\usepackage[accsupp]{axessibility}
\crefname{section}{Sec.}{Secs.}
\Crefname{section}{Section}{Sections}
\Crefname{table}{Table}{Tables}
\crefname{table}{Tab.}{Tabs.}

\def\wacvPaperID{57} 
\def\confName{WACV}
\def\confYear{2025}

\begin{document}
\thispagestyle{empty}
\title{Revisiting an Old Perspective\\Projection for Monocular 3D Morphable Models Regression}

\author{Toby Chong\\
TOEI Company\\
Tokyo, Japan\\
{\tt\small tob\_chong@toei.co.jp}
\and
Ryota Nakajima\\
TOEI Company\\
Tokyo, Japan\\
{\tt\small ryo\_nakajima@toei.co.jp}
}
\twocolumn[{%
\maketitle

\begin{center}
  \includegraphics[width=\linewidth]{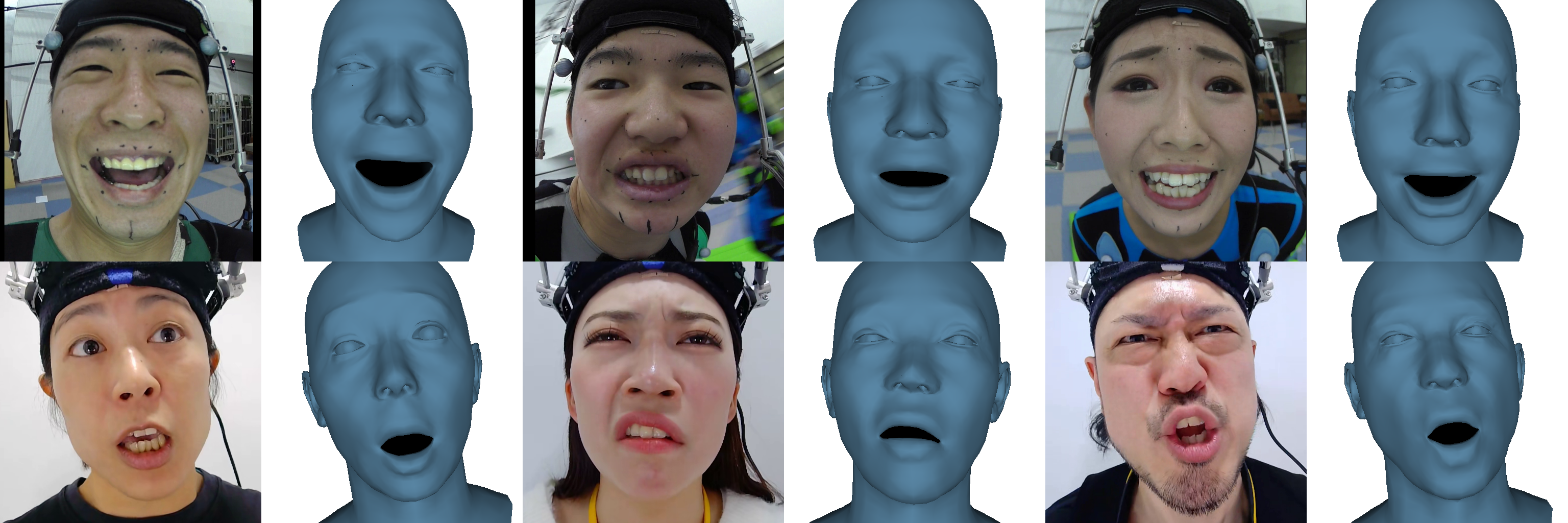}
  \captionof{figure}{
    We revisit perspective projection for 3DMM regression. We introduce a post hoc learnable parameter that is compatible with existing methods using orthogonal projection, to improve the reconstruction quality on close-up images. 
  }\label{fig:teaser}
\end{center}
\vspace{1em}
}]

\section{Abstract}
We introduce a novel camera model for monocular 3D Morphable Model (3DMM) regression methods that effectively captures the perspective distortion effect commonly seen in close-up facial images.

Fitting 3D morphable models to video is a key technique in content creation.
In particular, regression-based approaches have produced fast and accurate results by matching the rendered output of the morphable model to the target image. These methods typically achieve stable performance with orthographic projection, which eliminates the ambiguity between focal length and object distance. However, this simplification makes them unsuitable for close-up footage, such as that captured with head-mounted cameras.

We extend orthographic projection with a new shrinkage parameter, incorporating a pseudo-perspective effect while preserving the stability of the original projection.
We present several techniques that allow finetuning of existing models, and demonstrate the effectiveness of our modification through both quantitative and qualitative comparisons using a custom dataset recorded with head-mounted cameras.

\section{Introduction}
Fitting 3D morphable models (3DMMs) to videos traditionally required an optimization process to minimize the difference between the projected 3DMM and the 2D target image features in order to estimate the 3DMM parameters.  
With the development of deep learning techniques, regression-based approaches that predict the 3DMM parameters directly from image features have become more popular.  
By comparing the rendered image and the target image, it is possible to learn efficiently without ground truth annotations.  
However, these methods typically employ orthogonal projection to map the 3DMM onto image space, because it removes the ambiguity regarding focal length and distance from the camera.  
This projection method ignores the effect of perspective distortion, which leads to unintended artifacts such as smaller noses and levitated jawlines.

In this work, we propose a simple yet impactful camera model to capture the effect of perspective distortion by extending orthogonal projection with a \textbf{shrinkage parameter}, which controls the strength of perspective distortion. 
This parameter is regressed with a linear layer and is compatible with most architectures.  
We propose several fine-tuning techniques that allow us to adopt this parameter into existing methods that use orthogonal projection, including a per-dataset shrinkage prior and a masking technique to address the ambiguity of the nose and the face contour.
These processes enable converting existing models trained with orthogonal projection to the proposed camera model, using uncalibrated images.
We evaluate the effectiveness of our proposed modification through quantitative and qualitative comparisons on head-mounted camera footage.

In summary, our contributions are as follows:
\begin{itemize}[itemsep=0pt, topsep=0pt, parsep=0pt, partopsep=0pt, left=15pt]
    \item a new camera model which extends orthogonal projection to capture the effects of perspective distortion. 
    \item fine-tuning techniques to incorporate this shrinkage parameter into existing methods using orthogonal projection.
\end{itemize}

\section{Related work}\label{sec:background}

In this section, we review existing methods for fitting 3D morphable models (3DMMs) to images, focusing on unsupervised approaches that utilize rendering techniques. We will also discuss the limitations of orthogonal projection in these methods and introduce our proposed weak perspective projection modification.

\subsection{3D Morphable Models}\label{sec:background:3dmm}

There are a variety of 3D morphable models (3DMMs). 
Earlier approaches were based on linear combinations of 3D scans. Effectively, these methods perform a principal component analysis (PCA) on a set of 3D scans to create a statistical model that captures variations in shape and appearance across the dataset, such as the Surry Face Model~\cite{surrey_face_model} and the Basel Face Model~\cite{basel_face_model}. 
Combining a multitude of 3D scans, these models allow for the representation of a wide range of facial shapes and expressions using a relatively small number of parameters.

There is a substantial amount of effort to improve 3DMM, such as using neural representations~\cite{bouritsas2019neural, adapt3dmm} and nerf-based solutions~\cite{hong2021headnerf}. Although these methods are promising, the simplicity and compatibility with existing rendering engines of linear 3DMMs make them the de facto standard for many tasks involving 3DMMs.

One of the most commonly used 3DMM FLAME~\cite{FLAME:SiggraphAsia2017}, from which we borrow the notation below, expresses the morphable model $\mathcal{T}$ as follows:
\begin{equation}
 \mathcal{T}_P(\beta, \theta, \psi) = \bar{T} + B_S(\vec{\beta}; \mathcal{S}) + B_P(\vec{\theta}; \mathcal{P}) + B_E(\vec{\psi}; \mathcal{E})
\end{equation}
Where the final geometry $\mathcal{T}_P$ is a function of the mean shape $\bar{T}$, the shape basis $B_S$, the pose basis $B_P$, and the expression basis $B_E$. The parameters $\vec{\beta}$, $\vec{\theta}$, and $\vec{\psi}$ are the shape, pose, and expression coefficients, respectively. With $\mathcal{S}$ and $\mathcal{E}$ being the matrices of the basis of shape and expression, and $\mathcal{P}$ being the skinning function for different head poses. The shape basis captures the variations in the 3D shape of the face, while the pose basis captures the variations in the head's pose, and the expression basis captures the variations in facial expression. 

As such, theoretically, any face can be roughly represented with three sets of parameters ($\vec{\beta}$, $\vec{\theta}$, and $\vec{\psi}$), and methods to fit these parameters to facial portrait images or 3D scans have been a major focus of research in the field of computer vision and graphics.

\subsection{Fitting 3D Morphable Models}\label{sec:background:fitting_3dmm}
Fitting 3D Morphable Models (3DMMs) to images involves estimating the parameters of the model that best match the observed data. This process traditionally requires an optimization algorithm that minimizes the difference between the projected 3D model and the 2D image features.

Optimization-based methods often solve for the parameters of the 3DMM as well as camera parameters, such as camera pose and focal length, by minimizing a loss function that measures the discrepancy between the projected 3D model and the observed image features~\cite{blanz2003face, huber2015fitting, Booth_2017_CVPR, inequalityfit2020, gecer2021fast, giebenhain2025pixel3dmm}.

As more sophisticated image features and training methods have been developed, there has been a shift from optimization-based fitting to regression-based approaches. These methods leverage deep learning techniques to directly regress 3DMM parameters from image features~\cite{DECA:Siggraph2021, EMOCA:CVPR:2021, spectre2022visual, zielonka2022mica, hewitt2024look}. This process is often lightweight and enables possibilities such as end-to-end training, improved generalization to unseen data, and new applications like video-based fitting and real-time performance.

Unsupervised fitting methods such as DECA~\cite{DECA:Siggraph2021} and EMOCA~\cite{EMOCA:CVPR:2021} have shown promising results by leveraging rendering techniques to optimize the fitting process. These methods typically use differentiable rendering to compute gradients of the image features with respect to the 3DMM parameters, allowing for efficient optimization without requiring ground truth annotations.

SMIRK~\cite{SMIRK:CVPR:2024} uses a synthesis-based approach to create more accurate facial expressions by synthesizing images using a generator to regress the appearance of a portrait under different facial expressions. It has been shown that this approach can significantly improve the reconstruction quality of extreme expressions.

One common aspect of many of these regression-based methods is the use of orthogonal projection to map the 3D model onto the 2D image plane. While this simplifies the fitting process by ignoring perspective distortion, it also introduces unintended artifacts in the final results. In this work, we propose a backward-compatible modification to the orthogonal projection that better captures the perspective distortion effect, leading to improved reconstruction quality, especially for close-up images.

\section{Methods}\label{sec:methods}

\subsection{Projection Methods Used in 3D Morphable Models}\label{sec:background:projection_methods}
The fitting of 3DMMs to images typically involves projecting the 3D model onto a 2D image plane. The most common projection methods used in 3DMM fitting are perspective projection and orthogonal projection.

Perspective projection is more commonly used in traditional optimization-based 3DMM fitting~\cite{zielonka2022mica, hewitt2024look}, as it provides a more realistic representation of how 3D objects appear in 2D images. While more complex models exist, the most widely adopted form of perspective projection is based on the pinhole camera model, which simulates a camera with a single point of view.

In this section, we largely follow the notation from Booth et al.~\cite{Booth_2017_CVPR}.
Given a point $q$ in the canonical space of the 3DMM mesh, the point is first transformed by the camera parameters, which include rotation and translation, and can be expressed as:
\begin{equation} 
[v_x, v_y, v_z] = R_vq + [t_x, t_y, t_z]
\end{equation}

Then the perspective projection onto a 2D image plane can be expressed as:
\begin{equation}
\label{equ:persp_proj}
\begin{pmatrix}
u \\
v
\end{pmatrix}
=
\begin{pmatrix}
\frac{f_x \cdot v_x}{v_z} + c_x \\
\frac{f_y \cdot v_y}{v_z} + c_y
\end{pmatrix}
\end{equation}

Where $(u, v)$ are the pixel coordinates in the image, $(v_x, v_y, v_z)$ are the 3D coordinates of the point, $(f_x, f_y)$ are the focal lengths in the x and y directions, and $(c_x, c_y)$ are the coordinates of the principal point. This is often simplified by setting $f_x = f_y = f$ and $c_x = c_y = 0$, resulting in the following simplified equation:
\begin{equation}
\label{equ:persp_proj_simplified}
\begin{pmatrix}
u \\
v
\end{pmatrix}
=
\begin{pmatrix}
\frac{f \cdot v_x}{v_z} \\
\frac{f \cdot v_y}{v_z}
\end{pmatrix}
\end{equation}

Regressing the camera parameters requires estimating the 3D rotation $R$, 3D translation $t$, and focal length $f$, resulting in a total of seven parameters.

Nevertheless, regression-based methods often avoid using perspective projection. One of the earlier works that explored perspective projection~\cite{Booth_2017_CVPR} concluded that ``it is beneficial to keep the focal length constant in most cases, due to its ambiguity with $t_z$''. This is indeed true; a small face in an image could result either from the face being far from the camera or from the camera having a small focal length, with almost no way to distinguish between the two—except in one noticeable exception (Section~\ref{sec:background:orthogonal_missing}). 
Training a network to correctly predict both parameters is difficult, as the network would have to learn to compensate $t_z$ with $f$ (Section~\ref{sec:eval:ablation}).

Humans can often resolve this ambiguity using additional cues, such as the background in the image, but to the best of our knowledge, no existing method attempts to directly regress focal length from the image.

Instead, focal length is typically: 1.\ completely ignored, as in the case of orthogonal projection~\cite{EMOCA:CVPR:2021, DECA:Siggraph2021, SMIRK:CVPR:2024, zielonka2022mica}, 2.\ fixed to a constant value~\cite{deng2019accurate, tewari2017mofa}, or 3.\ estimated in post-processing via optimization~\cite{hewitt2024look, giebenhain2025pixel3dmm, wood2022dense}.

Orthogonal projection is often adopted to simplify the fitting process by ignoring the effects of perspective distortion. In practice, this is typically implemented as:

\begin{equation}
\label{equ:ortho_proj}
\begin{pmatrix}
u \\
v
\end{pmatrix}
=
\begin{pmatrix}
Sv_x\\ 
Sv_y
\end{pmatrix}
\end{equation}

Where $S$ is a scaling factor, effectively ignoring the depth information ($t_z$, $v_z$) of the 3D model. As a result, the regression of the camera parameters is reduced to estimating only the 3D rotation $R$, 2D translation components $t_x$ and $t_y$, and the scaling factor $S$, for a total of six parameters.

This eliminates the ambiguity between focal length and $t_z$, and has become the de facto standard in regression-based 3DMM fitting methods.

\subsection{What Orthogonal Projection Misses}\label{sec:background:orthogonal_missing}

\begin{figure*}[htbp]
    \centering
    \includegraphics[width=\textwidth]{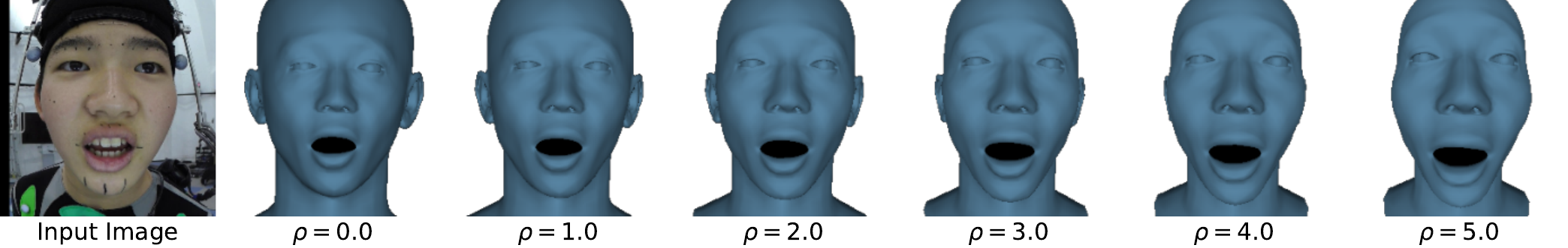}
    \caption{Visualization of the newly introduced shrinkage parameter $\rho$. We estimate the 3DMM and camera parameters using SMIRK~\cite{SMIRK:CVPR:2024}, which employs orthogonal projection ($\rho = 0$). We vary $\rho$ from 0.0 to 5.0 while holding all other parameters constant. Unlike perspective projection, which relies on the combination of $f$ and $t_z$ to control the shrinkage effect, $\rho$ isolates this effect and can therefore be incorporated into existing 3DMM regression methods via fine-tuning.}\label{fig:fov_effect_vis}
\end{figure*}

Our method is based on one key observation: existing 3DMM regression methods often fail to capture certain facial details accurately.
One prominent cue we observed is, quite literally, `on the nose'. When comparing the rendering of regressed outputs to in-the-wild images, as well as extreme close-up images filmed with a head-mounted camera (HMC), we noticed that the noses often appear significantly smaller than their actual size in the photo (Fig.\ref{fig:smirk_comparison}c, Fig.\ref{fig:fov_effect_vis}).

Given that the nose is the most protruding part of the face and often the closest to the camera in front-facing portraits (corresponding to a smaller $t_z$ value), it should appear larger under perspective projection than under orthogonal projection. Yet many existing methods reconstruct noses that are noticeably smaller than their real-world size in the image.

A similar effect can be observed around the contour of the face, where the upper `edge' of the 3DMM mesh tends to bend outward, effectively exaggerating the parietal region of the head. We refer to this as the `expanding brain' effect. This distortion is visible across all images generated by the baseline method in Figure~\ref{fig:hmc1m_comparison}.

Our method introduces a simple modification to the orthogonal projection used in existing 3DMM regression methods (Sec.\ref{sec:methods:orthogonal_projection}). This modification enables the network to capture perspective distortion effects through fine-tuning, even when initially trained with orthogonal projection. We evaluated this approach by fine-tuning SMIRK\cite{SMIRK:CVPR:2024} (Sec.\ref{sec:methods:finetuning}). Additionally, we introduced several minor components: a dataset of extreme close-up images captured with head-mounted cameras (Sec.\ref{sec:methods:dataset}), and a masking technique to address ambiguities around the nose and face contour during fine-tuning (Sec.~\ref{sec:methods:masking}).

\subsection{Pseudo Perspective Camera Model}\label{sec:methods:orthogonal_projection}
Our proposed camera model is an extension of the orthogonal projection method by adding a new shrinkage parameter $\rho$ that introduces the perspective effect.
{\small
\begin{equation} 
[v_x, v_y, v_z] = R_v q + [t_x, t_y, 0]
\end{equation}

\begin{equation}
\label{equ:ortho_fov}
\begin{pmatrix}
u \\
v
\end{pmatrix}
=
\begin{pmatrix}
S\frac{v_x}{1+\rho v_z}\\ 
S\frac{v_y}{1+\rho v_z}
\end{pmatrix}
\end{equation}
}

With $\rho = 0$, the projection is equivalent to orthogonal projection, and as $\rho$ increases, the projection becomes more perspective-like ($f \approx \frac{S}{\rho} \quad \text{when } \rho v_z \gg 1
$), It contains some very convenient properties. First, we can interpolate between orthogonal and perspective projection via a smooth transition, which can be optimized via backpropagation. This formulation also effectively isolates the shrinkage effect of perspective projection into its own parameter $\rho$, with minimal change in the overall projected size of the object.
As such, we train a network first with orthogonal projection, and then gradually introduce the perspective distortion effect.
We include a visualization of the effect of this modification in Figure~\ref{fig:fov_effect_vis}.

This new camera model can be related back to the perspective projection model by comparing the two projection equations Eq.\ref{equ:persp_proj_simplified} and Eq.\ref{equ:ortho_fov}. The shrinkage parameter $\rho$ can be expressed in terms of focal length $f$, scale $S$, and the final relative position of the object to the camera $v$:

{\small
\begin{align}
 f \frac{v_{x, y}}{v_z} &= S\frac{v_{x, y}}{1+\rho v_z} \\
    \rho &= \frac{S}{f} - \frac{1}{v_z}\label{equ:rho_fov_relation}
\end{align}
}
In perspective project the object is unscale ($S = 1$), hence the equation simplifies to $\rho = \frac{1}{f} - \frac{1}{v_z}$. 
An intuitive way to think about this is that $\rho$ represents the balance between the focal length and the distance to the object, which determines the strength of perspective distortion, while having minimal effect on the overall size of the object projected. 
Longer focal length (higher $f$) result in smaller perspective distortion and a smaller $\rho$ value; in contrast, object closer to the camera (lower $v_z$) experiences more perspective distortion and a higher $\rho$ value, ceteris paribus.


This modification can be incorporated into existing 3DMM regression methods that use orthogonal projection, such as EMOCA~\cite{EMOCA:CVPR:2021}, DECA~\cite{DECA:Siggraph2021}, and SMIRK~\cite{SMIRK:CVPR:2024}. 

\subsection{Head-Mounted Camera Dataset --- HMC1M}\label{sec:methods:dataset}
The datasets used to train the regression methods (EMOCA, DECA, and SMIRK) are a mix of studio-captured and in-the-wild images. These images are typically captured from a distance greater than 50 cm from the subject, so the perspective effect is not very pronounced.
We curated an internal dataset captured with a head-mounted camera system similar to Faceware\footnote{https://facewaretech.com/cameras}. A few sample images are shown throughout this paper in Fig.\ref{fig:teaser} and Fig.\ref{fig:hmc1m_comparison}.
The dataset contains 1 million images, with the camera typically ranging from 15 to 30 cm from approximately 200 professional actors. The focal length of the cameras is adjusted prior to capture such that the entire face fits within the frame with buffer at all sides, and varies slightly during filming due to lens adjustment and zooming.
The images were randomly sampled from in-house footage of acting sequences.
We fine-tune the models using HMC1M, MEAD~\cite{kaisiyuan2020mead}, FFHQ~\cite{ffhq}, and CelebA~\cite{liu2015celeba}.

\subsection{Finetuning Process}\label{sec:methods:finetuning}
In this section, we describe how we fine-tune SMIRK~\cite{SMIRK:CVPR:2024}.
For completeness, we first briefly discuss the training procedure of SMIRK, but we recommend the reader refer to the original paper for more details.

\subsubsection{SMIRK formulation}\label{sec:methods:smirk_training}
SMIRK~\cite{SMIRK:CVPR:2024} is a 3DMM regression method that uses a neural renderer to synthesize facial expressions. It builds on previous work such as EMOCA~\cite{EMOCA:CVPR:2021} and DECA~\cite{DECA:Siggraph2021}, but leverages a synthesized renderer to improve the reconstruction quality for extreme facial expression.

All these methods use encoders $E_\beta$, $E_\phi$, and $E_\theta$ to regress the camera pose $\beta$, 3DMM shape parameters $\phi$, and facial expression parameters $\theta$, respectively. 
In addition to the encoders, SMIRK and EMOCA also use a neural renderer $R$ that learns to reconstruct the original input image using the regressed info ($\beta, \phi, \theta$), and additional background information. This renderer enables the network to form a feedback loop, allowing it to adjust the regressed parameters to better match the input image.

\subsubsection{Learning the Shrinkage Effect with Uncalibrated Images}\label{sec:methods:unsupervised_learning}
Most head-mounted camera footages are uncalibrated, as the camera would often be adjusted to fit the actor's face within the frame as much as possible (changing focal length), and the distance between the camera and the actor's face would also vary due to head shape and movement (changing $v_z$). As such, it would be crucial to learn the shrinkage parameter $\rho$ in an unsupervised manner.

We add a single linear layer followed by a sigmoid activation to $E_\beta$, with a scaling hyperparameter $\rho_{\max}$ to limit the maximum value of the $\rho$ parameter. We initialize the linear layer with small weights and bias (both set to 0.01) to ensure that the initial projection is compatible with the pretrained network.

We learn to regress the new shrinkage parameter $\rho$ using a loosely set prior $\rho_{prior}$.
We randomly sample five images from HMC1M and generate the 3DMM and camera parameters using the pretrained network. We then manually adjust $\rho$ for each image such that the rendering result best matches the image visually.
Based on this, we empirically set $\rho_{prior} = 4.0$ for HMC1M and $\rho_{prior} = 0.0$ for all other datasets. The objective is to minimize the deviation of the predicted $\rho$ from the prior value using an L2 loss term $L_\rho = \lambda_{p}||\rho - \rho_{prior}||_2^2$, where $\lambda_{p} = 0.1$, and this loss function is added to the original training loss.

$\rho_{prior}$ can also be determined analytically if $f$ and $v_z$ are known, using Eq~\ref{equ:rho_fov_relation}. For HMC1M, $v_z$ approximates 15 to 30 cm, and the sensor size is typically around $1/2.3\,$" (width $\approx 0.455\text{cm}$). Lenses are adjusted such that the face (width $\approx 15\text{cm}$) occupies approximately half of the image width. The effective $f$ and $\rho$ can be calculated as follows:
{\small
\begin{align}
 f &= \frac{\text{sensor width} \cdot v_z}{\text{face width} \cdot 2} 
     \approx 0.227 \text{–} 0.455 \\
 \rho &= \frac{1}{f} - \frac{1}{v_z} \approx 2.16 \text{-} 4.34
\end{align}
}
\subsubsection{Masking the Ambigious}\label{sec:methods:masking}
As discussed in Sec.\ref{sec:background:orthogonal_missing}, the nose and the face contour often appear to be the most mismatched regions. To address this, we modified the masking technique used in SMIRK\cite{SMIRK:CVPR:2024} to specifically handle the ambiguity of the nose and face contour. During training, we follow SMIRK by masking out all pixels of the entire face, then add back one percent of the pixels as guidance for the renderer $R$. We then erode the mask so that pixels around the contour are not added back. Additionally, we apply a separate mask at the center of the face to remove pixels around the nose.

\section{Evaluation}

\subsection{Evaluation Settings}
We fine-tune the model starting from DECA, following mostly the same procedures as SMIRK. We train our model using all the datasets employed in SMIRK, as well as our own dataset discussed in Section~\ref{sec:methods:dataset}, which we refer to as `Ours'. For a fair comparison, we also retrain SMIRK on the same datasets and refer to this version as $smirk_{r}$. Additionally, we evaluate the pretrained official SMIRK model, referred to as $smirk_{p}$. Where applicable, we also evaluate our method on MICA~\cite{zielonka2022mica}.

\subsection{Quantitative Evaluation}\label{sec:eval:quantitative}
We quantitatively evaluate the reconstruction quality of our method using two criteria: \textbf{1.} whether it can correctly produce a mesh that, when projected, matches the image, and \textbf{2.} whether it better recovers the underlying facial geometry.

\subsubsection{2D Landmark Reconstruction}
To address the first question, we evaluate the sparse landmark reconstruction loss on the test sets of in-the-wild datasets, including MEAD~\cite{kaisiyuan2020mead}, CelebA~\cite{liu2015celeba}, and FFHQ~\cite{ffhq}. Additionally, we assess reconstruction quality on the HMC1M dataset.
In Table~\ref{table:landmark_reconstruction}, we report the reconstruction loss separately for the jawline and the rest of the facial landmarks to provide a clearer picture.

We observe that our method achieves the best reconstruction quality across different landmark regions on HMC1M, while $smirk_{r}$ (SMIRK~\cite{SMIRK:CVPR:2024} fine-tuned with HMC1M) produces the best result on the jawline for the MEAD dataset. Although this observation alone is not conclusive, we hypothesize that it is because the MEAD dataset exhibits subtle uncorrected perspective distortion, and training with the heavily distorted HMC1M dataset allows partial correction of this effect.

However, this comparison remains inconclusive - not only because the reported loss has been shown to correlate poorly with human perception of reconstruction quality~\cite{aldeneh2022towards, EMOCA:CVPR:2021, spectre2022visual}, but also because the landmark detector used to compute the reconstruction loss is often inaccurate for close-up images.
We did not evaluate the methods on CelebA and FFHQ because all images from these datasets were used to train the original SMIRK model.

\begin{table}[]
\scriptsize
\resizebox{\columnwidth}{!}{%
\begin{tabular}{lcll}
\multicolumn{1}{c}{\begin{tabular}[c]{@{}c@{}}Landmark\\ Region\end{tabular}} & Method & \multicolumn{1}{c}{MEAD} & \multicolumn{1}{c}{HMC1M} \\ \hline
\multicolumn{1}{l|}{} & $smirk_{p}$ & 7.941& 5.861 \\ \cline{2-4} 
\multicolumn{1}{l|}{Jaw line} & $smirk_{r}$& \textbf{7.004}&5.065\\ \cline{2-4} 
\multicolumn{1}{l|}{}& Ours&10.16&\textbf{4.646}\\ \hline
\multicolumn{1}{l|}{}& $smirk_{p}$ &2.763&3.236\\ \cline{2-4} 
\multicolumn{1}{l|}{Facial}& $smirk_{r}$  &2.058&1.737\\ \cline{2-4} 
\multicolumn{1}{l|}{}& Ours&\textbf{1.565}&\textbf{1.655}\\ \hline
\end{tabular}%
}
\caption{Landmark reconstruction loss on the MEAD and HMC1M datasets. We report the reconstruction loss for the jaw line and the rest of the facial landmarks separately. All numbers are multiplied by $10^4$ for readability.
}
\label{table:landmark_reconstruction}
\end{table}


\begin{figure}[htbp]
    \centering
    \includegraphics[width=\columnwidth]{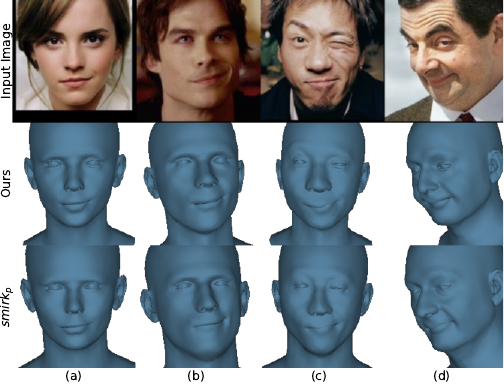}
    \caption{We compare our method with the pretrained SMIRK model, on the images used in the SMIRK paper. They provide similar reconstruction quality visibly. 
 }\label{fig:smirk_comparison}
\end{figure}

\subsubsection{3D-based Mesh Reconstruction}
As for the second question, we evaluate our method using the NoW dataset~\cite{RingNet:CVPR:2019}.
The NoW dataset is a large-scale collection of 3D scans of human faces in a neutral expression, accompanied by in-the-wild images.
In particular, we evaluated our method on the NoW neutral and selfie subsets. Since the NoW dataset does not provide ground-truth meshes for the selfie subset, we compare reconstruction quality by generating the neutral facial geometry using our method and $smirk_p$.
We first regress the full 3DMM and camera parameters using each method, then compute the neutral facial geometry by setting the expression and pose coefficients to zero. These final meshes are then compared with the 3D neutral face scans to calculate the reconstruction loss.
Table~\ref{table:now} reports the reconstruction loss for both the NoW neutral and selfie subsets.

While our method generally achieves better reconstruction quality than $smirk_p$, the difference is particularly significant on the NoW selfie subset. We attribute this to the fact that selfie images tend to exhibit stronger perspective distortion. Notably, our method performs better on the selfie subset than on the neutral subset, whereas $smirk_p$ shows the opposite trend. This aligns with our motivation: introducing perspective distortion should help with close-up images, such as selfies, where this effect is more pronounced.

That said, both $smirk_p$ and our method underperform compared to MICA~\cite{zielonka2022mica}. We attribute this to the fact that MICA is trained on a larger dataset of 3D scans specifically for the task of reconstructing neutral facial geometry from arbitrary expressions and viewing angles. In contrast, our method is optimized to produce a mesh that matches the input image, rather than the underlying neutral geometry.

\begin{table}[]
\resizebox{\columnwidth}{!}{%
\begin{tabular}{ccc}
Reconstruction Loss & NoW Neutral                 & NoW Selfie \\ \hline
\multicolumn{1}{c|}{$smirk_p$} & \multicolumn{1}{c|}{1.2563} & 1.2718     \\ \hline
\multicolumn{1}{c|}{Ours}  & \multicolumn{1}{c|}{1.2320} & 1.2143     \\ \hline
\multicolumn{1}{c|}{MICA~\cite{zielonka2022mica}}  & \multicolumn{1}{c|}{1.1162} & 1.1238     \\ \hline\end{tabular}%
}
\caption{Reconstruction loss on the NoW dataset. We report the reconstruction loss for the NoW neutral and selfie subset. 
}
\label{table:now}
\end{table}

\subsection{Qualitative Evaluation}
\subsubsection{Perception Study}
We randomly selected 100 images from each of the test sets of MEAD and HMC1M.
For each image, we generated mesh renderings using our method, $smirk_p$, and $smirk_r$. We then crowd-sourced responses via Amazon Mechanical Turk to evaluate which of the three reconstruction methods produced geometry that best matched the image. The display order of the renderings was randomized, and respondents were asked to select the rendering that best represented the image. Out of the 619 responses collected, 44.4\% (275) favored our method, 23.4\% (145) favored $smirk_p$, and 32.1\% (199) favored $smirk_r$.

\begin{figure*}
    \centering
    \includegraphics[width=0.9\textwidth]{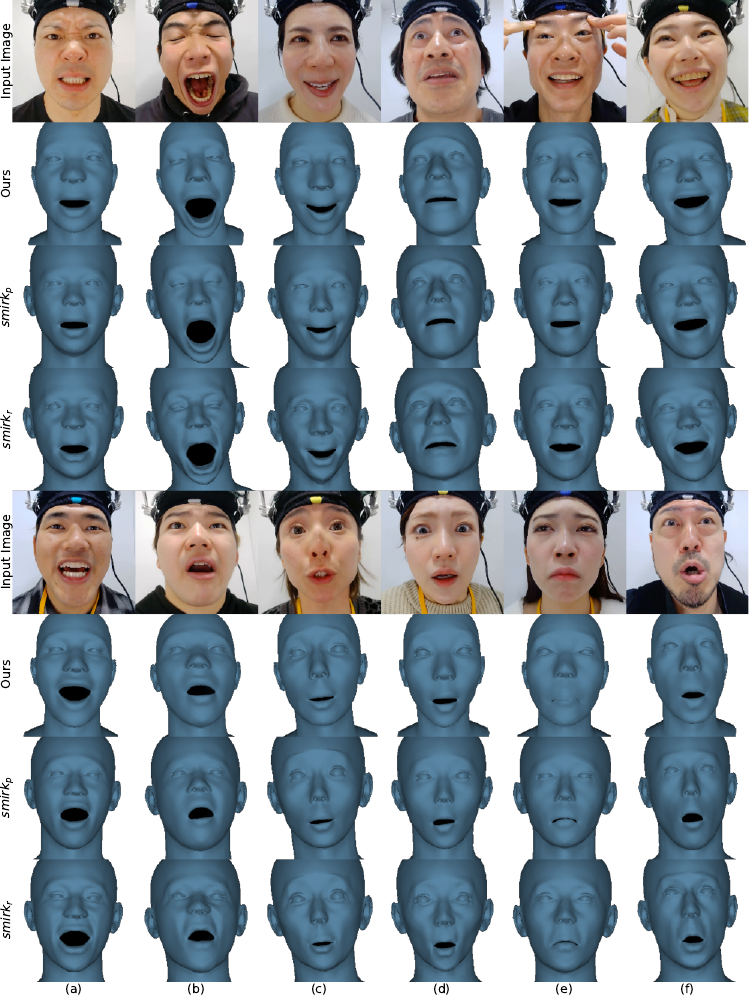}
    \caption{We compare our method with the pretrained and retrained versions of the SMIRK model, on the images from the HMC1M dataset. 
 }\label{fig:hmc1m_comparison}
\end{figure*}

\subsubsection{Visual Comparison}
We also provide visualizations comparing our method with $smirk_p$.
In Figure~\ref{fig:smirk_comparison}, we compare our method with $smirk_p$ using in-the-wild style teaser images from the SMIRK paper. While we observe minor improvements in specific regions—such as the nose in Fig.~\ref{fig:smirk_comparison}c—we do not find any significant difference in overall reconstruction quality. This is expected, as our method shares a largely identical network architecture and training protocol with SMIRK. 

Additionally, we include more visual comparisons using images from the HMC1M dataset (Fig.\ref{fig:hmc1m_comparison}).
In general, our method is more capable of reconstructing the geometry of close-up images. In the jaw region, it is clearly visible that $smirk_p$ produces unrealistic facial expressions—such as hollow cheeks—to compensate for perspective distortion (see Fig.\ref{fig:hmc1m_comparison}c, d). Our method also avoids the `expanding brain' issue seen in baseline outputs.


\section{Discussion}\label{sec:discussion}

\begin{table}[]
\centering

\resizebox{\columnwidth}{!}{%
\begin{tabular}{|c|c|c|c|c|}
\hline
        & HMC1M            & MEAD            & CelebA           & FFHQ            \\ \hline
$\rho$  & $2.95 \pm 0.96$ & $1.35 \pm 0.42$ & $0.69 \pm 0.31$ & $0.68 \pm 0.29$ \\ \hline
\end{tabular}%
}
\caption{Estimated $\rho$ values for all datasets used. We sampled and computed the average $\rho$ from 10k images from each of the datasets, under our camera model. HMC1M experiencing the highest distortion, and CelebA and FFHQ exhibiting minimal distortion.}\label{tab:rho}
\end{table}

\subsection{Limitations}
\label{sec:discussion:limitations}
Our motivation behind this work is to improve the reconstruction quality of monocular 3DMM regression methods for close-up images. Theoretically, our method can produce more accurate 3D reconstructions even for in-the-wild images. However, we did not observe a significant improvement in those cases. We suspect this is because most in-the-wild images are mostly orthogonal to begin with.
We measured and reported the average value of estimated $\rho$ from different datasets in Table~\ref{tab:rho}. 
HCM1M experiences the highest estimated $\rho$ value. This validates the need for a specialized camera model for handling extreme close-up images. CelebA and FFHQ both have very low $\rho$ values (little perspective distortion). This aligns with the fact that these datasets mostly consist of professionally taken images, which typically avoid extreme close-ups and are taken afar. MEAD has a moderate $\rho$ value, showing that it contains moderate perspective distortion. While we cannot find the documentation regarding the exact capturing setup (camera position, focal length), the capturing studio appears relatively small, and it would be reasonable to assume that the subjects are fairly close to the camera, which would explain the moderate $\rho$ value.
This is also likely the reason why our method achieves slightly better results on the MEAD dataset, as the baseline model trained with orthogonal projection would not be able to fully correct the distortion, and this highlights the importance of compensating for perspective distortion even for in-the-wild images.


\subsection{Training with full perspective model}
\label{sec:eval:ablation}

\begin{figure}[htbp]
    \centering
    \includegraphics[width=\columnwidth]{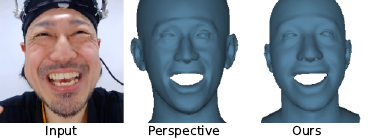}
    \caption{We finetuned the baseline SMIRK model with full perspective projection. The network would fail to properly capture the perspective distortion effect, and the results remain mostly orthogonal (large $f$ value). For comparsion, this converts to $\rho \approx 0.78$ in our camera model, while our method predicts a more visually similar result with $\rho=1.94$.
 }\label{fig:perspective_result}
\end{figure}

While early work such as Booth et.al.~\cite{Booth_2017_CVPR} concluded that learning to regress the focal length $f$ and distance to the object $t_z$ directly is difficult. As the reconstruction capability of neural networks improves, it is worth revisiting this question. As an alternative strategy, we also attempted to learn to regress $f$ and $t_z$ directly. 
First, we modified the projection method of SMIRK to predict $f$ and $t_z$ directly instead of $S$. This is, however, not successful, and the network loss would not stabilize as with orthogonal projection. We suspect it is due to the fact that the network would not be able to receive a stable enough gradient signal to adjust both $f$ and $t_z$ in opposite directions early in the training.
We also attempt to convert the pretrained SMIRK model to predict $f$ and $t_z$ via finetuning. We approximate the focal length with $f = S^{-1}$ and set up a reasonable $t_z$ value based on the training dataset. While this network is able to produce stable results, it failed to capture the perspective distortion effect, and the results remain mostly orthogonal (Fig~\ref{fig:perspective_result}). 
\section{Conclusion}\label{sec:conclusion}
In this work, we present a novel camera model by extending orthogonal projection method used in existing 3DMM regression approaches, and finetuning method for converting models trained with orthogonal projection to the new camera model. We show that our method effectively captures the shrinkage effect of perspective projection, which is crucial for accurately reconstructing facial geometry in close-up scenarios such as those captured by head-mounted cameras and selfies. 

Through extensive experiments, we demonstrate that our method outperforms existing approaches on the HMC1M dataset, which contains close-up facial images captured using head-mounted cameras. We also observe no significant performance degradation on more in-the-wild datasets, such as CelebA and FFHQ.
{\small
\bibliographystyle{ieee_fullname}
\bibliography{egbib}
}

\end{document}